 \documentclass[smallabstract,smallcaptions]{dccpaper}

\usepackage{epsfig}
\usepackage{amsmath}
\usepackage{amssymb}
\usepackage{color}
\usepackage{url}
\usepackage{pifont}

\newlength{\figurewidth}
\newlength{\smallfigurewidth}

\begin{document}

\title{\large
\textbf{L-STEC: Learned Video Compression with Long-term Spatio-Temporal Enhanced Context}
}

\author{%
Tiange Zhang$^{1,2,3}$, Zhimeng Huang$^{3,\ast}$, Xiandong Meng$^{2,\ast}$, Kai Zhang$^{4}$,\\ Zhipin Deng$^{4}$ and Siwei Ma$^{1,2,3,\ast}$\thanks{$^{\ast}$Corresponding authors: Zhimeng Huang (\protect\url{zmhuang@pku.edu.cn}), Siwei Ma (\protect\url{swma@pku.edu.cn}) and Xiandong Meng (\protect\url{mengxd@pcl.ac.cn}).}\\[0.5em]
{\small\begin{minipage}{\linewidth}\begin{center}
\begin{tabular}{c}
$^{1}$ Shenzhen Graduate School, Peking University, Shenzhen, China \\
$^{2}$Pengcheng Laboratory, Shenzhen, China \\
$^{3}$National Engineering Research Center for Visual Technology,\\ School of Computer Science, Peking University, Beijing, China \\
$^{4}$Bytedance Inc, San Diego, USA 
\end{tabular}
\end{center}\end{minipage}}
}

\maketitle
\thispagestyle{empty}

\begin{abstract}
Neural Video Compression has emerged in recent years, with condition-based frameworks outperforming traditional codecs. However, most existing methods rely solely on the previous frame's features to predict temporal context, leading to two critical issues. First, the short reference window misses long-term dependencies and fine texture details. Second, propagating only feature-level information accumulates errors over frames, causing prediction inaccuracies and loss of subtle textures. To address these, we propose the Long-term Spatio-Temporal Enhanced Context (L-STEC) method. We first extend the reference chain with LSTM to capture long-term dependencies. We then incorporate warped spatial context from the pixel domain, fusing spatio-temporal information through a multi-receptive field network to better preserve reference details. Experimental results show that L-STEC significantly improves compression by enriching contextual information, achieving 37.01\% bitrate savings in PSNR and 31.65\% in MS-SSIM compared to DCVC-TCM, outperforming both VTM-17.0 and DCVC-FM and establishing new state-of-the-art performance.
\end{abstract}

\Section{Introduction}

Video compression is essential for reducing the storage and transmission costs of video data, especially as high-definition content becomes increasingly prevalent. Traditional video compression methods, such as H.265/HEVC~\cite{sullivan2012overview} and H.266/VVC~\cite{bross2021overview}, have long formed the backbone of multimedia systems. However, these approaches have reached a performance ceiling, which limits further improvements and motivates the development of learning-based solutions. Neural Video Compression (NVC) addresses this limitation by replacing traditional codec modules with a fully neural framework, enabling end-to-end compression. When trained on large-scale video datasets, NVC effectively removes redundant information through estimation and prediction, achieving strong generalization and improved compression efficiency.

Among various NVC frameworks~\cite{jia2025emerging,SIP-20250056}, the condition-based paradigm has proven most effective and currently represents the state-of-the-art. Condition-based codecs typically rely on two main processes. The motion coding process estimates and compensates motion based on features from previous frames, capturing temporal dependencies. The context coding process leverages these temporal features to encode and decode the current frame more efficiently, forming the foundation of context-based compression.

Despite these advances, most existing NVC methods~\cite{li2021deep,sheng2022temporal,li2022hybrid,zhai2024hybrid,li2023neural,li2024neural,LIN2023126396,10222075} depend solely on the previous frame features to generate temporal context. This introduces two critical problems. First, the narrow reference window fails to capture long-term dependencies, which are essential for maintaining reconstruction quality over long video sequences. Second, propagating information purely in the feature domain easily accumulates errors: small inaccuracies in early frames amplify over time, quickly degrading prediction and erasing fine details. Relying solely on features cannot preserve subtle textures and high-frequency content, making pixel-domain information indispensable. Together, these limitations severely hinder compression performance and visual fidelity, highlighting the urgent need for long-term temporal and spatial context modeling that leverages both feature and pixel information.

To overcome these challenges, we propose a novel method named Long-term Spatio-Temporal Enhanced Context (L-STEC). Our approach addresses the limitations of short reference windows and error propagation in two key ways. First, we extend the reference chain using Long Short-Term Memory (LSTM) networks, enabling the model to capture long-term dependencies and enrich context across multiple frames. Second, we extract spatial context from warped pixel-domain features and fuse it with temporal context through a multi-receptive field network. This fusion retains more fine-grained information that conventional single-frame temporal approaches often lose.

By enriching contextual information for compression, L-STEC achieves superior performance. Our main contributions are summarized as follows:
\begin{itemize}
\item \textbf{Long-term Reference Chain Modeling:} We extend the reference chain using LSTM networks to capture long-term dependencies, improving prediction accuracy and supplementing long-term features.
\item \textbf{Spatio-Temporal Context Mining:} We extract spatial context from pixel-domain warped features and fuse it with temporal context via a multi-receptive field network, which enhances the retention of fine details.
\item \textbf{Experimental Validation:} Experiments show that L-STEC reduces bitrate by 37.01\% in PSNR and 31.65\% in MS-SSIM compared to DCVC-TCM and surpasses VTM-17.0 and DCVC-FM, establishing new state-of-the-art results in neural video compression.
\end{itemize}

\Section{Related Work}
Deep Video Compression (DVC)~\cite{9072487} is regarded as the pioneer in neural video compression, marking the beginning of a shift from traditional video codecs to end-to-end learned models. Video compression tasks primarily focus on removing temporal, spatial, and statistical redundancies. Based on inter-frame prediction strategies, video compression frameworks can be broadly categorized into four groups: 1) Methods based on 3D convolutions~\cite{habibian2019video}, which treat video sequences as three-dimensional spatio-temporal data for processing, 2) Probabilistic modeling approaches~\cite{yang2020learning}, which use networks to model probability distributions and predict video content, 3) Implicit representation based methods~\cite{chen2021nerv}, which use neural networks to directly overfit video segments and transmit the learned parameters and 4) Motion estimation and compensation based methods~\cite{9072487}, which focus on the prediction of motion between frames and compensate for these motions to reduce redundancy. Among these, condition-based paradigms belonging to the fourth category have achieved the most significant improvements, with the DCVC~\cite{li2021deep} series serving as a landmark in this area.

The DCVC framework has since evolved with many architectural and optimization advances. Various modifications have been proposed to address key challenges in neural video compression. These include improving motion estimation accuracy~\cite{10416688}, controlling bitrate more efficiently~\cite{li2024neural}, enhancing context modeling~\cite{tang2025neural}, optimizing entropy models~\cite{li2022hybrid} and refining training methods.

\begin{figure}[t!]
\centering
\includegraphics[width=0.93\textwidth]{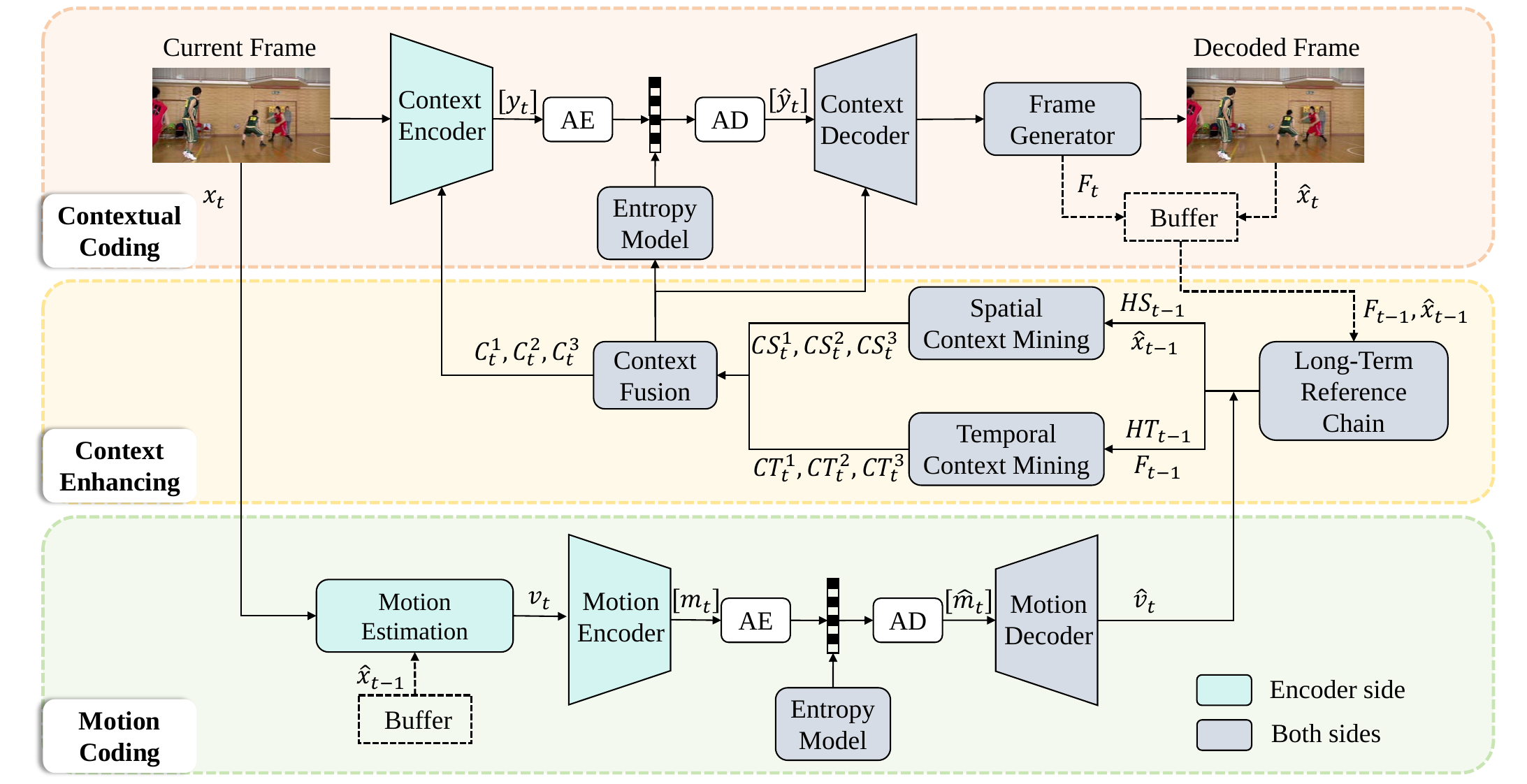}
\caption{Overview of the proposed L-STEC framework. The Long-Term Reference Chain extracts $HT_{t-1}$ and $HS_{t-1}$ from the previous frame $\hat{x}_{t-1}$ and its features $F_{t-1}$, which are fused into spatio-temporal enhanced contexts $C_t^{1}, C_t^{2}, C_t^{3}$ for guiding the coding of $x_t$.}
\label{fig:framework}
\vspace{-1em}
\end{figure}

Context optimization has become a major research focus in neural video compression. Several approaches have explored better capturing temporal dependencies between video frames. In DCVC~\cite{li2021deep}, context is initially modeled by warping motion vectors and features from reconstructed frames to capture temporal dependencies. DCVC-TCM~\cite{sheng2022temporal} expands this by introducing multi-scale context and a finer refinement module for improved accuracy. DCVC-DC~\cite{li2023neural} introduces diversity into temporal context by predicting multiple offset groups, extracting more varied and robust contextual information. DCMVC~\cite{tang2025neural} incorporates context compensation to modulate the propagated temporal context from the reference feature. HLGC~\cite{zhai2024hybrid} designs a local-global context enhancement module to fully explore the local-global information of previous reconstructed signals. However, while these methods advance temporal context optimization, they focus solely on the temporal domain without addressing spatial context optimization. Thus, the potential benefits of optimizing spatial context remain largely unexplored in these frameworks.

\Section{Proposed Method}
\subsection*{\textbf{Overview}}
The framework of our proposed L-STEC is shown in Fig.~\ref{fig:framework}, which consists of three main components: motion coding, long-term spatio-temporal enhanced context mining, and context coding.

In the motion coding phase, the motion vector \( v_t \) is estimated using SpyNet between the current frame \( x_t \) and the previous reconstructed frame \( \hat{x}_{t-1} \). The flow is encoded into a latent representation, quantized as \( m_t \), and compressed by entropy coding. After transmission, the flow is decoded and reconstructed as \( \hat{v}_t \).

In the long-term spatio-temporal enhanced context mining phase, the long-term reference chain enhances feature extraction by exploring both spatial and temporal dependencies from the previous reconstructed frame \( \hat{x}_{t-1} \), the associated feature \( F_{t-1} \), and the LSTM hidden cell, which stores previous reference frames and their features. This process outputs hidden features \( {HT}_{t-1} \) and \( {HS}_{t-1} \). The flow, frame, and features are then processed through spatial context mining and temporal context mining modules. These context features are fused to compensate for the motion and produce the final enhanced context.

In the context coding phase, the current frame \( x_t \) is encoded into a latent representation, quantized as \( y_t \), and compressed via entropy coding. After decoding and dequantization, the frame generator reconstructs the frame \( \hat{x}_t \), guided by the enhanced context. Finally, \( \hat{x}_t \) and its features are stored in the buffer for subsequent frames.

\subsection*{\textbf{Long-term Reference Chain Modeling}}
\begin{figure}[t!]
\centering
\includegraphics[width=\textwidth]{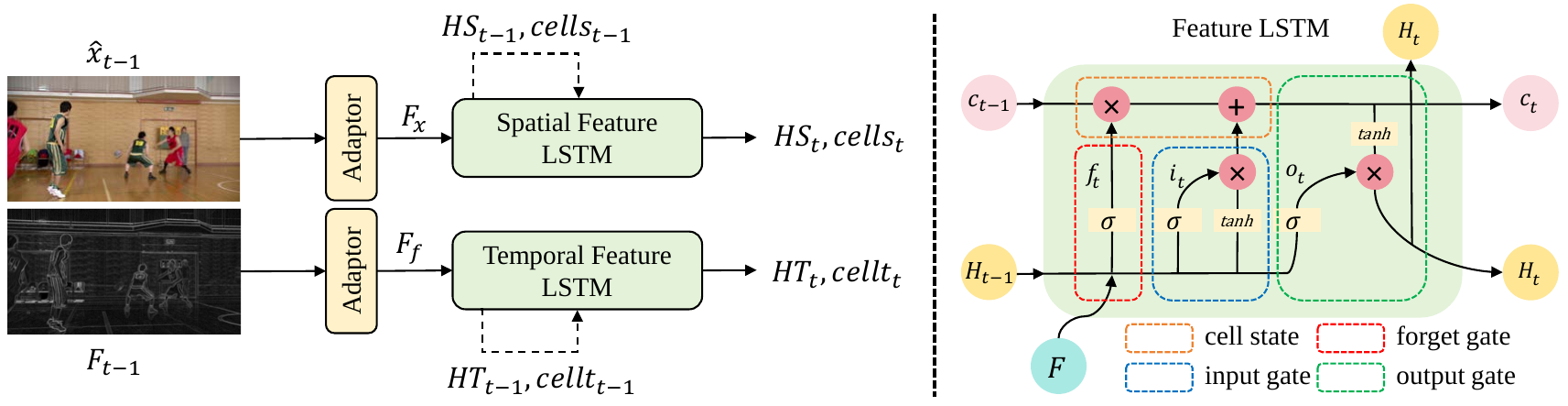}
\caption{Illustration of the long-term reference chain module, where $\hat{x}_{t-1}$ and $F_{t-1}$ are processed through LSTMs with gates to update cell states and produce hidden spatial features $HS_t$ and temporal features $HT_t$.}
\label{fig:chain}
\end{figure}

This chain is designed to capture and maintain spatial and temporal reference information over extended frames, as shown in Fig.~\ref{fig:chain}. The module is divided into two key components: long-term spatial feature extraction and temporal feature extraction. The current frame's reconstructed feature \( F_{t-1} \) and the previous frame \( \hat{x}_{t-1} \) are fed into two separate CNN-based adaptors for spatial and temporal feature processing, respectively:
\begin{align}
F_x &= \text{Adaptor}(\hat{x}_{t-1}), \\ F_f &= \text{Adaptor}(F_{t-1}).
\end{align}

The spatial feature adaptor first processes \( \hat{x}_{t-1} \) to extract spatial information. The resulting features are then fed into an LSTM network, which outputs the hidden spatial feature \( HS_t \) along with its cell state \( cells_t \). In parallel, the temporal feature adaptor takes the previous frame feature \( F_{t-1} \) to capture temporal dependencies. These features are subsequently passed through another LSTM, producing the hidden temporal feature \( HT_t \) and the corresponding temporal cell state \( cellt_t \):
\begin{align}
HS_t, \, cells_t &= \text{SpatialFeatureLSTM}(F_x, HS_{t-1}, cells_{t-1}), \\
HT_t, \, cellt_t &= \text{TemporalFeatureLSTM}(F_f, HT_{t-1}, cellt_{t-1}).
\end{align}

\begin{figure}[t!]
\centering
\includegraphics[width=\textwidth]{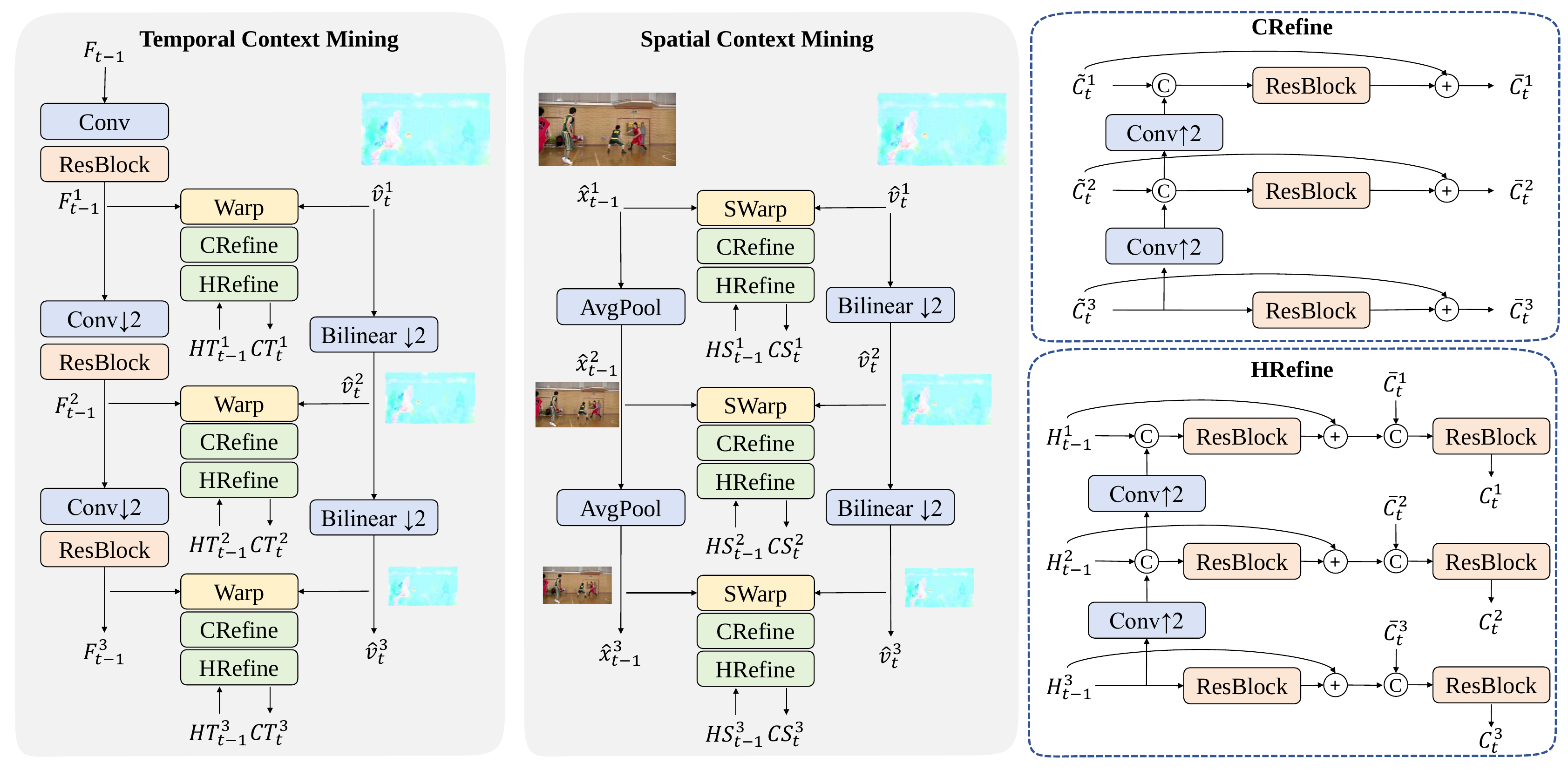}
\caption{Illustration of the spatio-temporal context mining module. Previous frame features $F_{t-1}$ and $\hat{x}_{t-1}$ are first warped with motion. Then they are refined by CRefine for context and HRefine for hidden information, yielding temporal and spatial contexts.}
\vspace{-1em}
\label{fig:mining}
\end{figure}

The LSTM operations can be described by the following set of equations:
\begin{align}
i_t &= \sigma(W_i \times [H_{t-1}, F_t] + b_i), \\
f_t &= \sigma(W_f \times [H_{t-1}, F_t] + b_f), \\
o_t &= \sigma(W_o \times [H_{t-1}, F_t] + b_o), \\
c_t &= f_t \odot c_{t-1} + i_t \odot \tanh(W_c \times [H_{t-1}, F_t] + b_c), \\
h_t &= o_t \odot \tanh(c_t).
\end{align}
where $i_t, f_t, o_t$ are the input, forget, and output gates, respectively. $c_t$ is the cell state at time step $t$, and $H_t$ is the hidden state $HS_t$ or $HT_t$, which is used as the output of the LSTM at each time step. $F_t$ denotes the input feature, either $F_x$ or $F_f$. The weights $W$ and biases $b$ are learned during training.

By maintaining separate spatial and temporal context modeling, we ensure that the long-term dependencies from both domains are captured effectively. Additionally, the LSTM’s ability to retain long-term memory allows the framework to propagate useful context over time.

\subsection*{\textbf{Spatio-Temporal Context Mining and Fusion}}
\begin{figure}
\centering
\includegraphics[width=\textwidth]{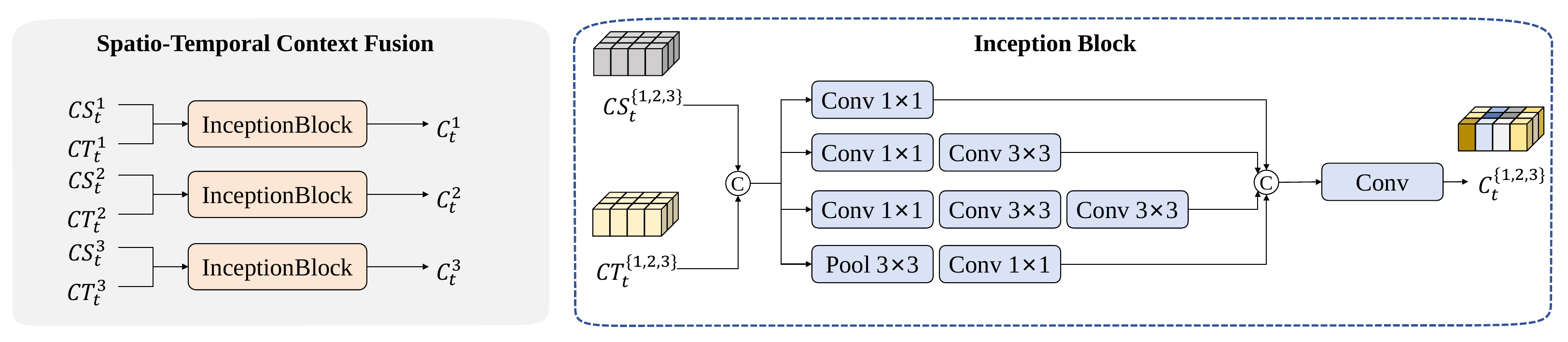}
\caption{Illustration of the spatio-temporal context fusion module. Temporal context $CT_t^{1,2,3}$ and spatial context $CS_t^{1,2,3}$ are fused to produce unified enhanced contexts $C_t^{1,2,3}$.}
\vspace{-1em}
\label{fig:fusion}
\end{figure}

This module plays a crucial role in extracting and combining relevant contextual information for enhanced context generation, as shown in Fig.~\ref{fig:mining} and Fig.~\ref{fig:fusion}. The spatio-temporal context mining consists of three key components: temporal context mining, spatial context mining, and spatio-temporal context fusion.

\textbf{Temporal Context Mining:}  In this phase, the optical flow \( v_t \) undergoes bilinear downsampling, and the previous frame feature \( F_{t-1} \) is processed through convolutional downsampling to obtain aligned features at three different scales. Then, the warped features are refined through the CRefine and HRefine modules. The CRefine module refines the context information, while the HRefine module refines the hidden information. The two refining modules work in tandem to integrate both context and hidden features, providing a more comprehensive representation of the temporal context: 
\begin{align}
CT_{t}^{1,2,3} = HRefine(CRefine(Warp(F_{t-1}^{1,2,3}, \hat{v}_t^{1,2,3})), HT_{t-1}^{1,2,3}).
\end{align}

\textbf{Spatial Context Mining:} This process works with the current reconstructed frame \( \hat{x}_{t-1} \), where it extracts spatial information. The frame \( \hat{x}_{t-1} \) undergoes a similar procedure: first passed through a spatial downsampling module avgpool, then processed using the SWarp (Spatial Warp), CRefine, and HRefine modules for further refinement:
\begin{align}
CS_{t}^{1,2,3} = HRefine(CRefine(SWarp(\hat{x}_{t-1}^{1,2,3}, \hat{v}_t^{1,2,3})), HS_{t-1}^{1,2,3}).
\end{align}

\textbf{Spatio-Temporal Context Fusion:} In this phase, the temporal context \( CT_t \) and spatial context \( CS_t \) are fused to create a unified context representation that captures both the spatial and temporal dependencies. The fusion process is done using the Inception Block, which combines multiple convolution layers with different kernel sizes and pool layers to effectively combine spatial and temporal information:
\begin{align}
C_t^{1,2,3} = \text{InceptionBlock}(CS_t^{1,2,3}, CT_t^{1,2,3}).
\end{align}

\Section{Experiments}

\subsection*{\textbf{Experimental Setup}}
\textbf{Datasets.} For training, we only use the Vimeo-90K dataset~\cite{xue2019video}, which is widely used in neural video compression. During training, the original video frames with a resolution of 448 × 256 are randomly cropped into 256 × 256 patches. For evaluation, we assess the performance of the proposed method on several commonly applied test datasets, including the UVG dataset~\cite{mercat2020uvg}, MCL-JCV dataset~\cite{wang2016mcl}, and the JCT-VC sequences~\cite{sullivan2012overview}. Since the test datasets are not in RGB format, we follow DCVC-DC~\cite{li2023neural} and apply BT.709 color space conversion for proper evaluation.

\noindent \textbf{Loss Function:} The loss function is designed to optimize the rate-distortion trade-off, defined as:
\begin{equation}
L_t = w_t \cdot \lambda \cdot d(x_t, \hat{x}_t) + r([m_t]) + r([y_t]),
\end{equation} where \( d(x_t, \hat{x}_t) \) is the distortion metric (MSE or 1-MS-SSIM), $w_t$ refers to hierarchical training weight and \( r([m_t]) \), \( r([y_t]) \) represent the bit rate for encoding the motion and context features. 


The cascaded loss strategy is used to reduce error propagation, with fine-tuning performed in the final epochs. The total loss for a video clip is averaged over \( T \) frames:
\begin{equation}
L_T = \frac{1}{T} \sum_{t=1}^{T} L_t.
\end{equation}
\begin{table}[t!]
\centering
\caption{BD-Rate (\%) comparison in RGB color space (BT.709) measured with PSNR.}
\vspace{-0.5em}
\resizebox{0.95\textwidth}{!}{ 
\renewcommand{\arraystretch}{1.1} 
\begin{tabular}{c|ccccc|c}  
\hline
\textbf{Method} & \textbf{HEVC B} & \textbf{HEVC C} & \textbf{HEVC D} & \textbf{UVG} & \textbf{MCL-JCV} & \textbf{Average} \\
\hline
DCVC-TCM  & 0.00 & 0.00 & 0.00 & 0.00 & 0.00 & 0.00  \\
HM-16.25 & -1.74 & -17.79 & -2.36 & -8.15 & -7.71 & -7.55  \\
VTM-17.0 & -28.81 & -40.10 & -26.98 & -31.58 & -34.03 & -32.30  \\
DCVC-HEM & -23.73 & -29.23 & -27.53 & -26.89 & -26.23 & -26.73  \\
DCVC-FM & -27.82 & -40.53 & \textbf{-42.12} & -33.76 & -31.19 & -35.08  \\
Our method & \textbf{-33.88} & \textbf{-40.86} & -40.46 & \textbf{-34.42} & \textbf{-35.43} & \textbf{-37.01} \\
\hline
\end{tabular}}
\vspace{-1em}
\label{table:bdrate}
\end{table}

\begin{table}[t!]
\centering
\caption{BD-Rate (\%) comparison in RGB color space (BT.709) measured with MS-SSIM.}
\vspace{0.5em}
\resizebox{0.95\textwidth}{!}{  
\renewcommand{\arraystretch}{1.1} 
\begin{tabular}{c|ccccc|c}
\hline
\textbf{Method} & \textbf{HEVC B} & \textbf{HEVC C} & \textbf{HEVC D} & \textbf{UVG} & \textbf{MCL-JCV} & \textbf{Average} \\
\hline
DCVC-TCM  & 0.00 & 0.00 & 0.00 & 0.00 & 0.00 & 0.00  \\
HM-16.25 & 53.38 & 56.93 & 87.56 & 23.67 & 52.28 & 54.76   \\
VTM-17.0 & 12.91 & 16.26 & 41.26 & 1.12 & 13.51 & 17.01  \\
DCVC-HEM & -19.77 & -26.49 & -27.03 & -16.86 & -20.74 & -22.18  \\
DCVC-DC & -27.80 & \textbf{-38.56} & \textbf{-41.56} & -23.37 & -26.88 & -31.63 \\
Our method & \textbf{-29.21} & -37.42 & -39.02 & \textbf{-24.63} & \textbf{-27.95} & \textbf{-31.65} \\
\hline
\end{tabular}}
\vspace{-1em}
\label{table:bdratessim}
\end{table}

\noindent \textbf{Implementation Details:} We develop our method by integrating network components from DCVC-DC~\cite{li2023neural} with the framework design of DCVC-TCM~\cite{sheng2022temporal}, 
using 4 $\lambda$ values (85, 170, 380, 840) to control the rate-distortion trade-off. The hierarchical weight \( w_t \) is set as (0.5, 1.2, 0.5, 0.9) for the 4 frames. The model is implemented in PyTorch with AdamW optimization and a batch size of 4. The training first optimizes PSNR for 33 epochs, followed by 5 epochs of fine-tuning. This setup ensures efficient rate-distortion balance and model stability.
For MS-SSIM optimization, the model is fine-tuned for 3 additional epochs.
\begin{figure}[t!]
\centering
\includegraphics[width=\textwidth]{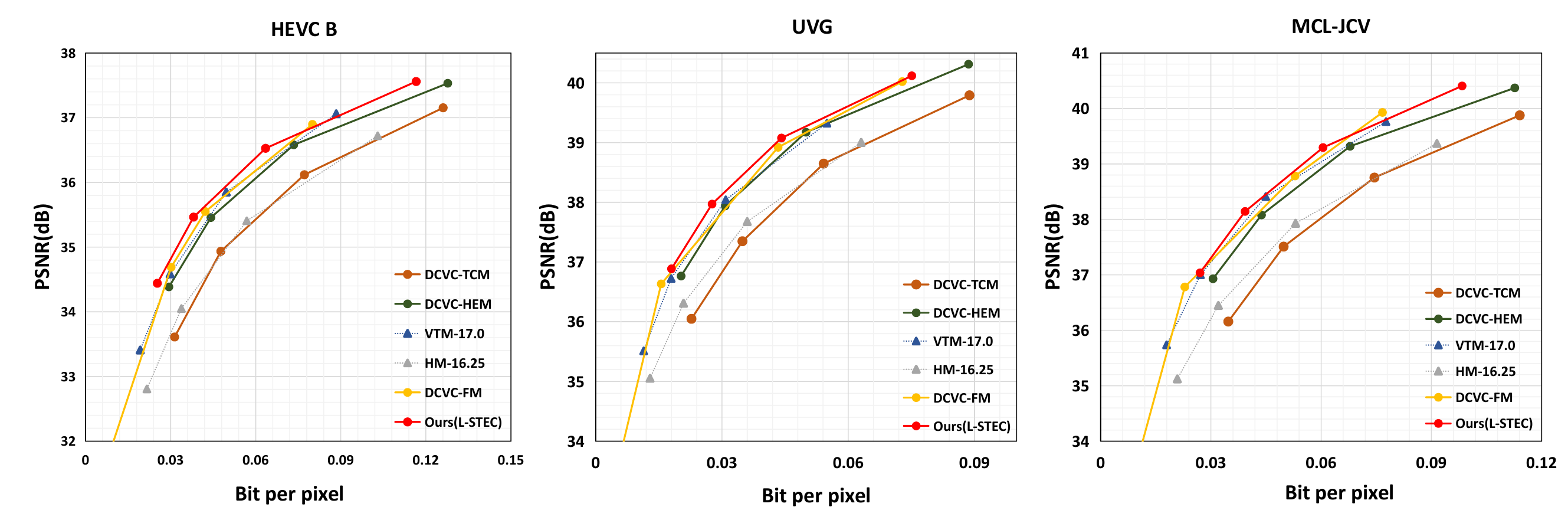} 
\caption{R-D curve of the proposed and compared methods, evaluated on the HEVC ClassB, UVG, and MCL-JCV datasets in RGB color space (BT.709) with PSNR.}
\label{fig:rd_performance}
\vspace{-1em}
\end{figure}

\vspace{-0.5em}
\subsection*{\textbf{Experimental Results}}
We evaluate the proposed method on HEVC ClassB, C, D, UVG, and MCL-JCV datasets using an intra period of 32 and testing the first 96 frames. The I-frame model is consistent with that in DCVC-DC. We pad the video frames to multiples of 16 and use real encoding and decoding processes. We compare our method with state-of-the-art approaches, including DCVC-TCM, HM-16.25, VTM-17.0, DCVC-HEM, DCVC-DC and DCVC-FM. Table.~\ref{table:bdrate} reports the BD-Rate (\%) comparisons measured by PSNR, where lower values indicate better compression performance. In terms of PSNR, our method consistently outperforms all compared methods, surpassing VTM-17.0 and achieving superior results compared to DCVC-FM. Specifically, we achieve an average BD-Rate reduction of 37.01\% relative to DCVC-TCM. The R-D curves in Fig.~\ref{fig:rd_performance} further confirm that our approach provides improved compression efficiency and enhanced reconstruction quality.

Table.~\ref{table:bdratessim} presents the BD-Rate comparisons measured by MS-SSIM, which focuses on structural similarity and shows strong correlation with MOS. Our method attains an average BD-Rate saving of 31.65\% compared to DCVC-TCM and also outperforms DCVC-DC, indicating better preservation of high-frequency details and visual textures.

\begin{figure}[t!]
\centering
\includegraphics[width=0.9\textwidth]{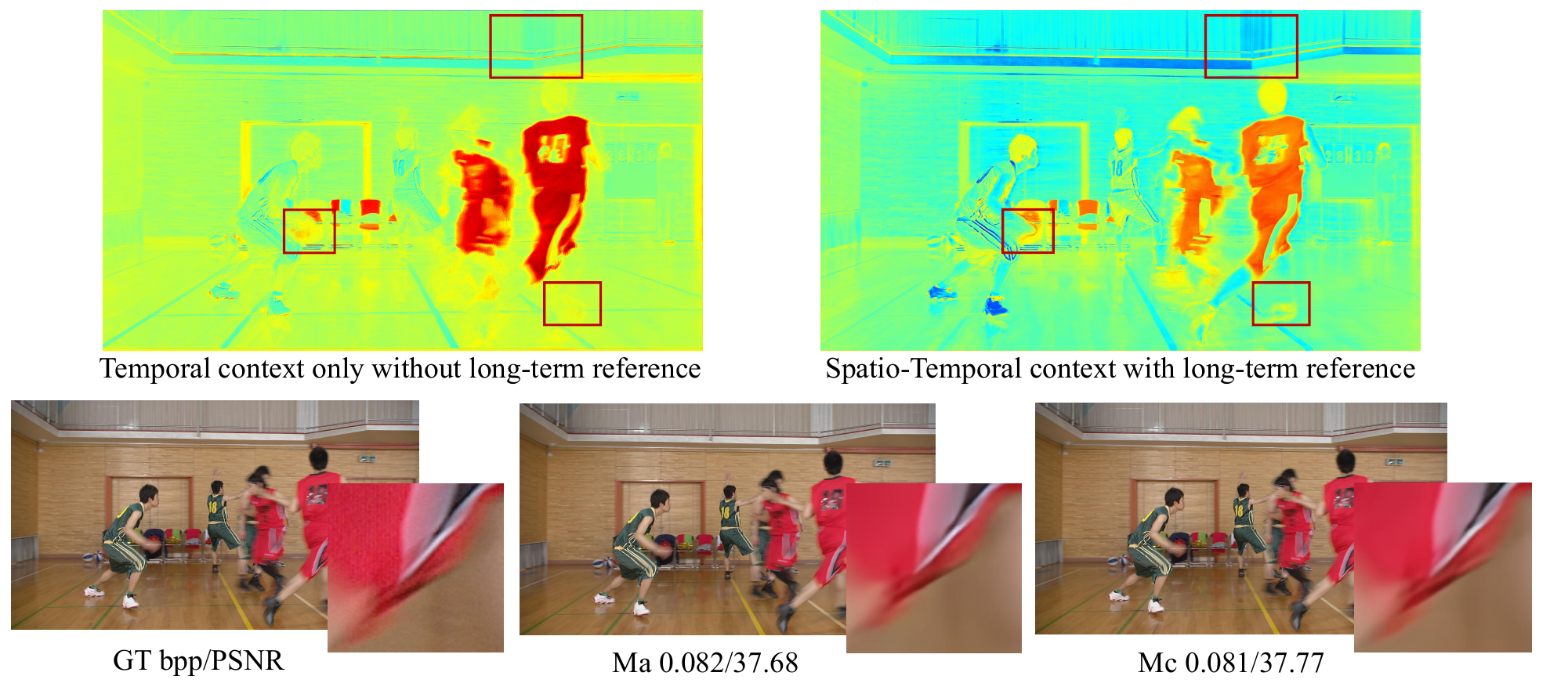}  
\caption{Visualization results on one frame of HEVC ClassB basketball sequence at the highest scale for the Ma and Mc models.}
\label{fig:visualization_results}
\vspace{-1em}
\end{figure}

\vspace{-0.5em}
\subsection*{\textbf{Ablation Study}}

To evaluate the impact of the key components in our method, we conduct an ablation study on the HEVC ClassB dataset in RGB color space, using PSNR as the evaluation metric. As summarized in Table.~\ref{ablation_study}, Model Ma serves as the baseline with only temporal context mining, while Mb further introduces a long-term reference chain, achieving a 5.6\% reduction in BD-Rate. Mc additionally integrates spatio-temporal context mining and fusion, leading to another 2.9\% BD-Rate improvement. 

To better illustrate these effects, we visualize the context feature on the HEVC ClassB basketball sequence in Fig.~\ref{fig:visualization_results}. Specifically, we average the 48 output channels to generate a single heatmap, where the color transition from blue to green to red denotes increasing attention to local details. The comparison demonstrates that the spatio-temporal context with long-term reference yields stronger contrast between moving and static regions, and preserves more long-term fine-grained details. This richer feature representation substantially enhances the reconstruction performance.

\begin{table}[t!]
\renewcommand{\arraystretch}{1.1}
\centering
\caption{Ablation study on the main components.}
\vspace{-0.7em}
\resizebox{0.7\textwidth}{!}{ 
\begin{tabular}{c|c c c }
\hline
\textbf{Method} & \textbf{Ma} & \textbf{Mb} & \textbf{Mc} \\
\hline
Long-term reference chain & \ding{55} & \ding{51} & \ding{51} \\
Temporal Context Mining & \ding{51} & \ding{51} & \ding{55} \\
Spatio-Temporal Context Mining and Fusion & \ding{55} & \ding{55} & \ding{51} \\
\textbf{BD-Rate(\%)} & \textbf{0.0} & \textbf{-5.6} & \textbf{-8.5}  \\
\hline
\end{tabular}}
\vspace{-1em}
\label{ablation_study}
\end{table}

\Section{Conclusion}
\vspace{-0.5em}
This paper introduced the Long-term Spatio-Temporal Enhanced Context (L-STEC) method, addressing key limitations in existing neural video compression frameworks. L-STEC effectively captures long-term dependencies and enhances detail retention by using LSTM networks to extend the reference chain and incorporating warped spatial context. Experimental results demonstrate that L-STEC achieves significant bitrate reductions, clearly outperforming current state-of-the-art methods. This work highlights the importance of optimizing spatio-temporal context and opens avenues for further research into more efficient, long-term strategies.

\Section{Acknowledgement}
\vspace{-0.3em}
This work was supported in part by the Key Research \& Development Program of Peng Cheng Laboratory under grant PCL2024A02 and in part by Natural Science Foundation of China No. 62088102.

\Section{References}
\bibliographystyle{IEEEbib}
\bibliography{refs}

\end{document}